\documentclass[letterpaper, 10 pt, conference]{ieeeconf}  

\IEEEoverridecommandlockouts                              

\usepackage{graphicx} 
\usepackage{amsmath} 
\usepackage{amssymb}  
\usepackage{hyperref}
\usepackage{booktabs}

\renewcommand{\thefootnote}{}

\begin{document}

\title{\LARGE \bf
From Bird's-Eye to Street View: Crafting Diverse and Condition-Aligned Images with Latent Diffusion Model
}

\author{Xiaojie Xu, Tianshuo Xu, Fulong Ma and Yingcong Chen}

\maketitle

\renewcommand{\thefootnote}{}
\footnotetext{The authors Xiaojie Xu, Tianshuo Xu and Fulong Ma are with The Hong Kong University of Science and Technology(Guangzhou), Nansha District, Guangzhou, Guangdong, China. \texttt{\{xxu763, txu647, fmaaf\}@connect.hkust-gz.edu.cn}

The corresponding author Yingcong Chen is with The Hong Kong University of Science and Technology(Guangzhou), Nansha District, Guangzhou, Guangdong, China and The Hong Kong University of Science and Technology, Clear Water Bay, Kowloon, Hong Kong. \texttt{yingcongchen@ust.hk}
}

\begin{abstract}

We explore Bird’s-Eye View (BEV) generation, converting a BEV map into its corresponding multi-view street images. Valued for its unified spatial representation aiding multi-sensor fusion, BEV is pivotal for various autonomous driving applications. Creating accurate street-view images from BEV maps is essential for portraying complex traffic scenarios and enhancing driving algorithms. Concurrently, diffusion-based conditional image generation models have demonstrated remarkable outcomes, adept at producing diverse, high-quality, and condition-aligned results. Nonetheless, the training of these models demands substantial data and computational resources. Hence, exploring methods to fine-tune these advanced models, like Stable Diffusion, for specific conditional generation tasks emerges as a promising avenue. In this paper, we introduce a practical framework for generating images from a BEV layout. Our approach comprises two main components: the Neural View Transformation and the Street Image Generation. The Neural View Transformation phase converts the BEV map into aligned multi-view semantic segmentation maps by learning the shape correspondence between the BEV and perspective views. Subsequently, the Street Image Generation phase utilizes these segmentations as a condition to guide a fine-tuned latent diffusion model. This finetuning process ensures both view and style consistency. Our model leverages the generative capacity of large pretrained diffusion models within traffic contexts, effectively yielding diverse and condition-coherent street view images.

\end{abstract}

\section{INTRODUCTION}

The emerging era of autonomous driving hinges on the adoption of sophisticated technologies and representations to ensure optimal navigation and decision-making. Among these, the bird’s-eye view (BEV) holds a unique position. By offering a top-down, map-like representation, the BEV provides invaluable insights into the immediate environment, capturing pertinent obstacles and hazards. 

While BEV perception \cite{philion2020lift,zhou2022cross,li2022bevformer} has been a focal point in recent studies, promising to bridge the transformation between street-level views and overhead perspectives, BEV generation—specifically the synthesis of realistic street-view images from a predefined BEV semantic layout—offers untapped potential.

At its core, BEV generation \cite{swerdlow2023street} translates a semantic layout, which captures a traffic scenario, into tangible street-view images. This translation facilitates an enhanced visualization of traffic scenarios in a real-world setting, making the abstract more accessible. One of the most compelling applications of BEV generation is the intuitive interface it offers for traffic scene visualization and modification. BEV generation allows human operators and system designers to modify a layout effortlessly, producing corresponding street-view images via generative models. This not only streamlines the training of autonomous systems but also serves as an effective testing and validation tool.

\begin{figure}[t]
\centering
\includegraphics[width=\linewidth]{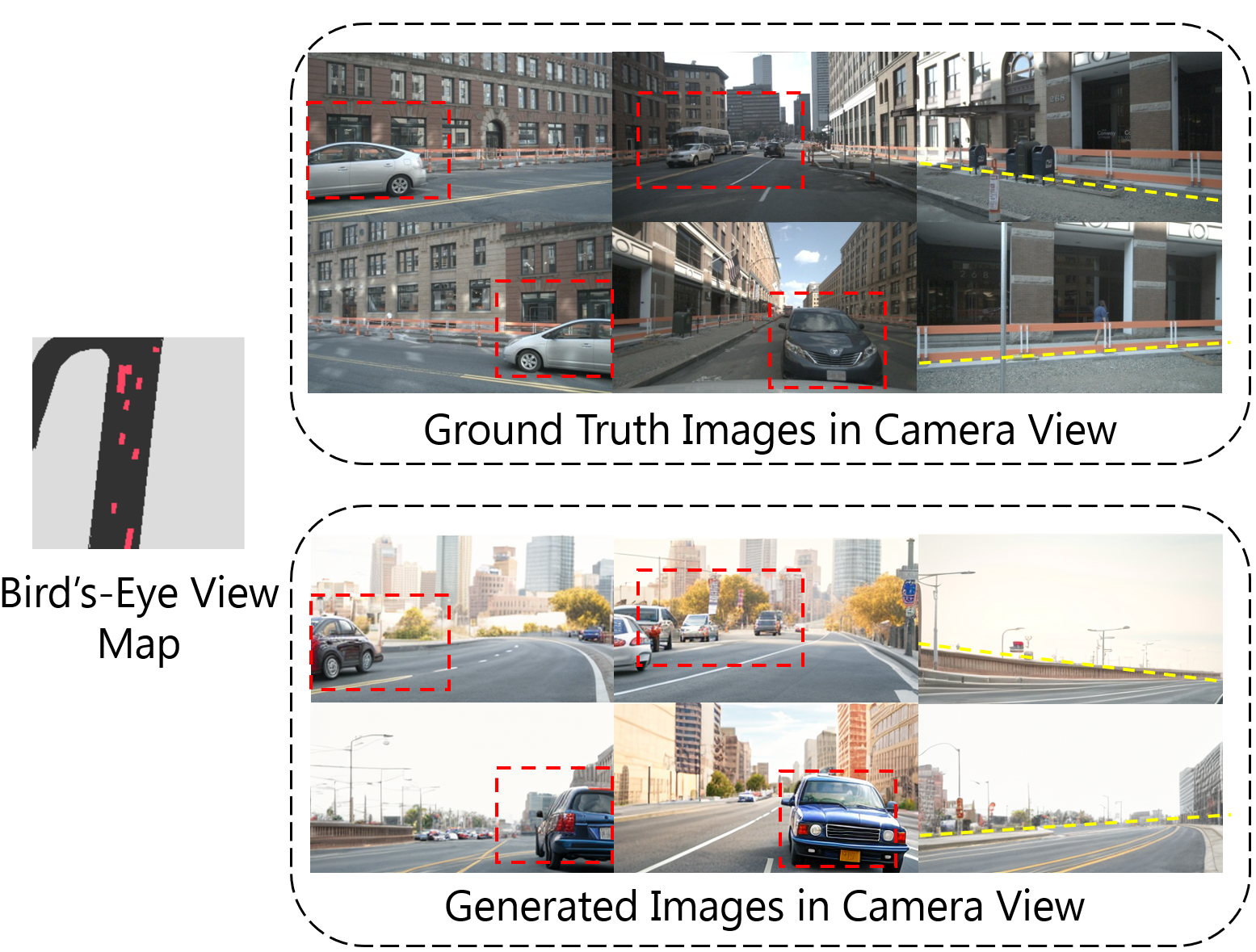}
\caption{From a bird's-eye view semantic map, our framework is capable of generating high-quality and varied camera view images. In terms of map elements, our results closely match the ground truth images. The red boxes (seen in the left four images) represent vehicles, while the yellow lines (in the right two images) delineate road contours.}
\label{teaser}
\end{figure}

BEVGen \cite{swerdlow2023street} represents a pioneering effort in addressing the BEV generation problem. Within its BEV representations, map components are bifurcated into two categories: vehicles and roads. The model employs an autoregressive transformer \cite{vaswani2017attention} with a spatial attention design to comprehend the relationship between camera and map perspectives. While BEVGen sets a baseline by generating multi-view images consistent with its map perspective, it doesn't consistently ensure condition coherence due to its implicit encoding mechanism.

In contrast to the previous method, our proposed method disentangles the view transformation and image generation processes. The view transformation phase focuses on learning the shape correspondence between map and camera perspectives. Here, to project the BEV map onto camera views using camera parameters, we assign height from a prior distribution to each BEV map segment. Using this projection as a preliminary estimate, a convolutional network is employed for shape refinement, achieving a more precise camera view segmentation. This refined segmentation acts as the conditional information for the image generator. For image synthesis, we resort to a conditional latent diffusion model \cite{rombach2022high}, chosen for its standout performance in conditional image generation tasks. Initially trained on diverse datasets, the diffusion model is fine-tuned using our driving scene imagery. Notably, the fine-tuning procedure encode camera viewpoint explicitly, ensuring that various views yield plausible outcomes (e.g., reasonable orientation of vehicles and roads). Leveraging precise transformed segmentation as condition and the generative ability of the diffusion model, our framework delivers high-quality, diverse, and condition-coherent results.

Our contributions are summarized as the following:
\begin{itemize}
    \item We develop a novel framework for street-view image generation from a BEV layout, leveraging a large, pretrained latent diffusion model. This encompasses view transformation, street-view adaptation, and conditional generation.
    \item We explore the methodology of encoding viewpoint for multi-view images and incorporating them into generative diffusion models, through which our method can produce diverse and flexible scenes that match the desired view and layout.
    \item We investigate the potential of utilizing large generative models for the task of BEV image generation and conduct a thorough comparison with other methods that are trained from scratch. Our method is efficient and effective, achieving high-quality and diverse results. Our experimental results demonstrate that our approach outperforms or matches existing methods in terms of visual quality and condition consistency.
\end{itemize}

\section{RELATED WORK}

\textbf{Conditional image generation:} The field of conditional image generation has seen notable advancements recently, with models predominantly conditioned on text \cite{reddy2021dall,saharia2022photorealistic} or speech \cite{chen2017deep} inputs. Varied formats, such as class conditions \cite{brock2018large}, sketches \cite{isola2017image}, style \cite{gatys2016image}, and distinct human poses \cite{ma2017pose}, can convey the envisaged image specifications. Furthermore, several scholars have explored methodologies with high level representations, including generating images from semantic masks \cite{park2019semantic} or translating intricate constructs like scene graphs \cite{dhamo2020semantic} and bounding boxes \cite{li2021image} into equivalent semantic masks. Diverging from these mentioned paradigms, our emphasis lies on the bird's-eye view map. Though akin to a semantic segmentation map, it offers a perspective distinct from the resulting image, which is seldom explored in earlier studies.

\textbf{Image diffusion models:} Originally proposed by Sohl-Dickstein et al. \cite{sohl2015deep}, Image diffusion models have found recent applications in image generation \cite{dhariwal2021diffusion}. The Latent Diffusion Models(LDM) \cite{rombach2022high} execute diffusion in the latent image space \cite{esser2021taming}, optimizing computational efficiency. Text-to-image diffusion models, by encoding textual inputs into latent vectors using pretrained language models like CLIP \cite{radford2021learning}, set new benchmarks in image generation. Glide \cite{nichol2021glide} stands out as a text-driven diffusion model for both image creation and editing. Stable Diffusion scales up the concept of latent diffusion \cite{rombach2022high}, and Imagen \cite{saharia2022photorealistic} takes a distinct approach by diffusing pixels through a pyramid structure, bypassing latent imagery. We employ Stable Diffusion as our foundational pretrained model. Through fine-tuning, we adapt it to various viewpoints and driving scenes.

\textbf{BEV perception and generation:}
Recent growth in large 3D datasets in autonomous driving \cite{caesar2020nuscenes,sun2020scalability,wilson2023argoverse} has propelled studies on map-view perception. Given the disparity between the coordinate frames of inputs and outputs, this domain poses challenges. While inputs derive from calibrated cameras, outputs are rasterized onto a map. A prevalent method assumes a mostly planar scene, simplifying image-to-map transformations via homography \cite{sengupta2012automatic}. However, this can create artifacts for dynamic entities like vehicles. As a solution, some studies \cite{wang2019parametric,schulter2018learning} utilize depth and semantic maps to present objects in BEV. Alternatively, other methods \cite{zhou2022cross,li2022bevformer} bypass explicit geometric modeling to generate map-view predictions directly from images. 

As its counterpart, generating from a BEV map layout remains relatively unexplored. BEVGen \cite{swerdlow2023street} pioneered this domain, employing an auto-regressive transformer to encode the connection between image and BEV representations. In contrast to BEVGen, our approach leverages a large, pretrained diffusion model as the backbone and finetunes it using driving scene images.

\begin{figure*}[t]
\centering
\includegraphics[width=0.9\linewidth]{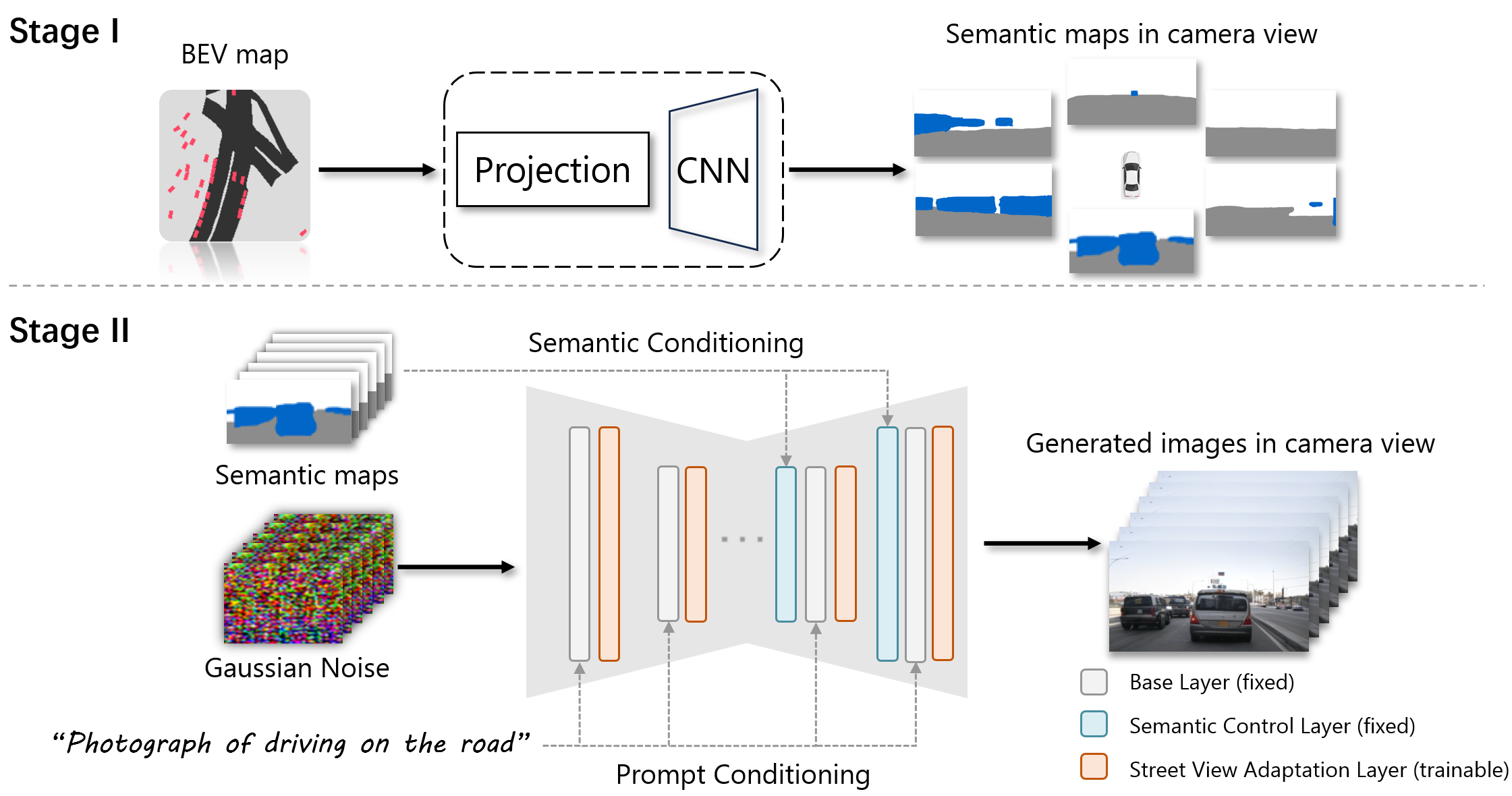}
\caption{Our two-staged pipeline. Initially, a BEV map is projected and refined to produce semantic maps from the camera's perspective. These semantic maps, paired with the prompt, are then fed into a pretrained U-Net for iterative denoising. We've incorporated street-view adaptation layers into the network to ensure style and viewpoint alignment.}
\label{pipeline}
\end{figure*}

\section{METHOD}

The objective of BEV generation is to generate multiple camera-view images from a semantic BEV layout. Earlier studies have represented the BEV layout in either rasterized \cite{zhou2022cross} or vectorized forms \cite{liu2023vectormapnet}. In this work, we favor the rasterized representation due to its aptness for creating from projections of 3D bounding boxes onto local street maps \cite{zhou2022cross}, or directly from driving simulation frameworks \cite{li2022metadrive}. Consequently, the BEV layout is denoted by $B\in \mathbb{R}^{H_b \times W_b \times c}$, where $c$ represents the number of map element categories, such as vehicles and roads. 

Given the BEV map $B$ and $n$ camera views $(K_i,R_i,t_i)_{i=1}^n$, where $K_i,R_i,t_i$ denotes the intrinsics, extrinsics rotation and extrinsics translation of the $i_{th}$ camera, our goal is to generate $n$ corresponding images in camera view $\mathcal{I}=\{\mathbf{I}^i\in\mathbb{R}^{H\times W\times 3}\mid i=1,...,n\}$. 

As depicted in Fig. ~\ref{pipeline}, our pipeline operates in two stages. Initially, the BEV's semantic information is projected into the camera view leveraging camera parameters, under a height assumption. This shape is subsequently refined using a CNN. In the succeeding stage, a pre-trained UNet undertakes the backward diffusion process \cite{rombach2022high}, where Gaussian noise is progressively eliminated. This UNet receives the polished semantic information coupled with the prompt as conditioning inputs. Furthermore, to ensure accurate viewpoints across various camera perspectives, we fine-tune the network.

\subsection{Stage I: Neural View Transformation} 

Taking inspiration from \cite{saha2022translating}, we treat the BEV-to-camera view transformation as an image translation task, where the input and output share a pronounced spatial correspondence. We decompose this transformation into two phases: initial setup using camera parameters and shape refinement via a neural network.

\textbf{Initial projection with camera parameters:}
For any world coordinate $X\in\mathbb{R}^{3}$, the perspective transformation describes its corresponding image coordinate $x\in\mathbb{R}^{3}$ in the view of the $i_{th}$ camera by

\begin{equation}\label{eq1}
x= K_iR_i(X-t_i)
\end{equation}
in homogeneous coordinates.

The lack of precise height data renders the world coordinates of BEV map data ambiguous, necessitating height estimation. While Inverse Perspective Mapping (IPM) techniques \cite{mallot1991inverse} operate under the premise of a flat ground, this assumption can introduce distortions for objects of varied heights, such as buildings and vehicles. Given our focus on roads and vehicles, we retain this simplified assumption for roads.

For vehicles, we posit that their height adheres to a predetermined distribution. Practically speaking, each vehicle on the BEV map is allocated a height randomly sampled from $\mathcal U(1.5, 2)$, offering a plausible initial height approximation. With the estimated heights for roads and vehicles in place, the BEV map is projected into camera views using Equ. \ref{eq1}, given the camera parameters.

\textbf{Shape refinement network:}
Through height estimation and projection, we obtain preliminary semantic maps in the camera view. Nonetheless, this simplistic initialization fails to preserve the intricate shapes of map elements accurately. Given that vehicles are rendered on the BEV map using their true 3D bounding boxes as described in \cite{zhou2022cross}, our projection approach results in the vehicle appearing as a cube from the camera's viewpoint. Hence, a shape-refinement post-processing step is imperative.

\begin{figure}[t]
\centering
\includegraphics[width=0.9\linewidth]{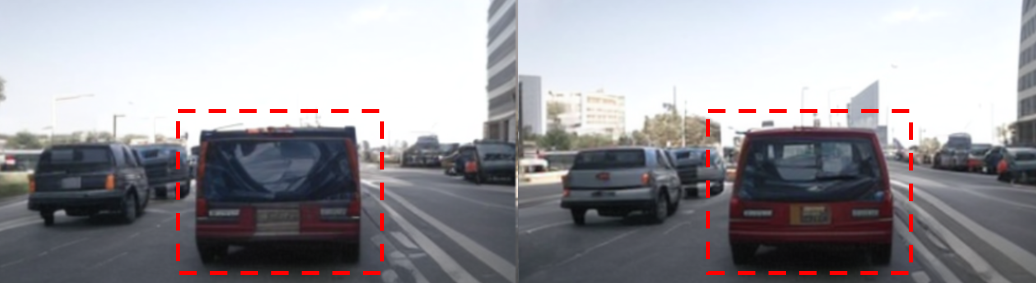}
\caption{The impact of shape refinement on the final image generation is evident. Without refinement, the resulting image (left) resembles a cube. In contrast, the refined version (right) exhibits a more natural form.}
\label{refine}
\end{figure}

The initial projection yields a low-resolution estimate. To address this, we employ an enhanced UNet architecture with residual connections \cite{chaurasia2017linknet}. This network bridges the shape discrepancy between the estimated and the true semantic maps. Functioning as an upsampling module, it outputs high-resolution semantic maps with finer geometry. These refined maps subsequently serve as conditional inputs to the image generator. The contribution of this network to the final image generation outcome is illustrated in Fig. ~\ref{refine}.

\begin{figure*}[t]
\centering
\includegraphics[width=0.75\linewidth]{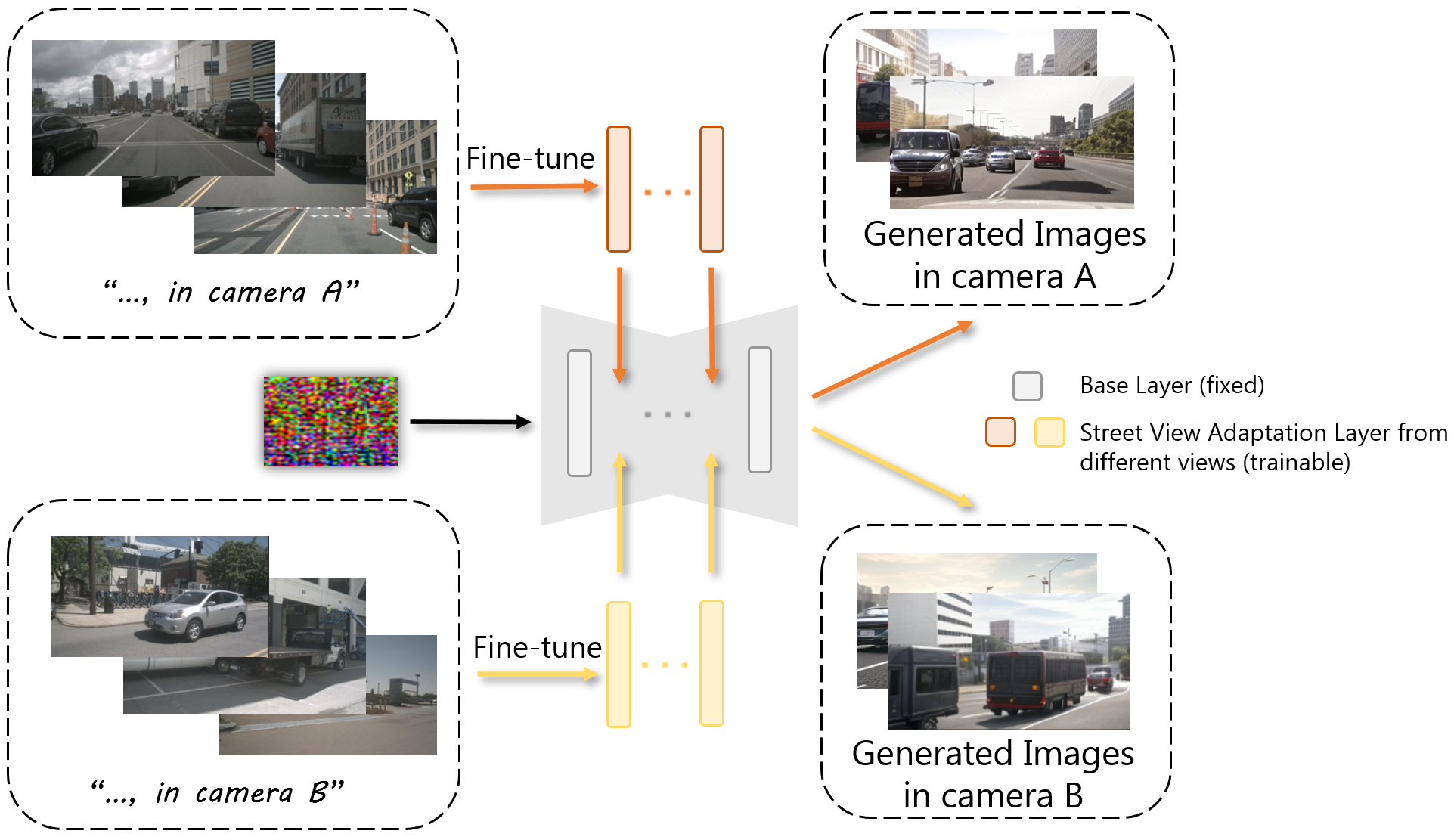}
\caption{We incorporate viewpoints into our foundational diffusion model by integrating specific views into the text prompts, resulting in distinct View Adaptation Layers. During sampling from the model, we can generate images from a designated camera by invoking its learned novel prompt. }
\label{view}
\end{figure*}

\subsection{Stage II: Street Image Generation}

We utilize Stable Diffusion, which is a strong pretrained image generator based on latent diffusion \cite{rombach2022high} framework, as our generative backbone. In this section we discuss how the conditional generation mechanism works and how to adapt the large pretrained model to our driving domain.

\textbf{Conditional generation with latent diffusion model:} Diffusion models can be conceptualized as a uniformly weighted sequence of denoising autoencoders, given by $\begin{aligned}\epsilon_\theta(x_t,t);t=1\ldots T\end{aligned}$, These autoencoders aim to predict a denoised version of their input $x_t$, where $x_t$ represents a noisy variant of the original input $x$. This leads to the following objective:
\begin{equation}\label{eq2}
    \begin{aligned}L_{DM}=\mathbb{E}_{x,\epsilon\sim\mathcal{N}(0,1),t}\left[\|\epsilon-\epsilon_\theta(x_t,t)\|_2^2\right]\end{aligned}
\end{equation}
with $t$ uniformly sampled from $\{1,\ldots,T\}$. 

As a large text-to-image diffusion model, latent diffusion introduces CLIP \cite{radford2021learning} encoder $\mathcal{T}_{\theta}$ that projects the text prompt $y$ to an intermediate representation $\mathcal{T}_{\theta}(y)$, which is then mapped to the intermediate layers of the UNet via a cross-attention layer implementing 
\begin{equation}
\text{Attention}(Q,K,V)=\text{softmax}\left(\frac{QK^T}{\sqrt{d}}\right)\cdot V
\end{equation}
, with $Q=W_Q^{(i)}\cdot\varphi_i(z_t),K=W_K^{(i)}\cdot\tau_\theta(y),V=W_V^{(i)}\cdot\tau_\theta(y)$.
In this context, $\varphi_i(z_t)$ symbolizes a flattened representation of the U-Net at an intermediate stage.

Our generation task encompasses more than just using the prompt as conditional information. The semantic data transformed from the BEV map serves as a superior control mechanism, given that the resultant image should align spatially with these semantic maps in pixel space. This necessitates a more precise conditioning mechanism for our objective.

Drawing inspiration from ControlNet \cite{zhang2023adding}, which employs zero convolution and a trainable duplicate of the original neural network, our approach manipulates the input conditions of neural network blocks. This strategy allows for a more nuanced control over the entire neural network's behavior. We integrate the pretrained ControlNet layers, designed for semantic segmentation, into our architecture (as depicted in Fig. ~\ref{pipeline}). These layers act as conditioning controllers for the image generation process. Even though these semantic control layers were trained on a broader dataset \cite{zhou2017scene}, they exhibit robust generalization capabilities in our driving scenarios.

\textbf{Street-view adaption:} Our street view adaptation module serves a dual purpose. Firstly, it emulates the driving scene's image style found in our dataset \cite{caesar2020nuscenes}. Secondly, it encapsulates the viewpoints associated with various cameras.

While fine-tuning the diffusion model using Equ. \ref{eq2} aids in capturing a realistic style specific to driving scenarios, it's crucial to remember that street scenes, when viewed from different camera perspectives, can vary significantly. For instance, when viewed through our front camera, a vehicle directly ahead should align with our car's driving direction. In contrast, the same vehicle observed from a side camera would appear at an angle. Likewise, driveable areas typically extend more prominently when viewed from the front and rear cameras but appear more constrained from the side angles.

\begin{figure*}[t]
\centering
\includegraphics[width=\linewidth]{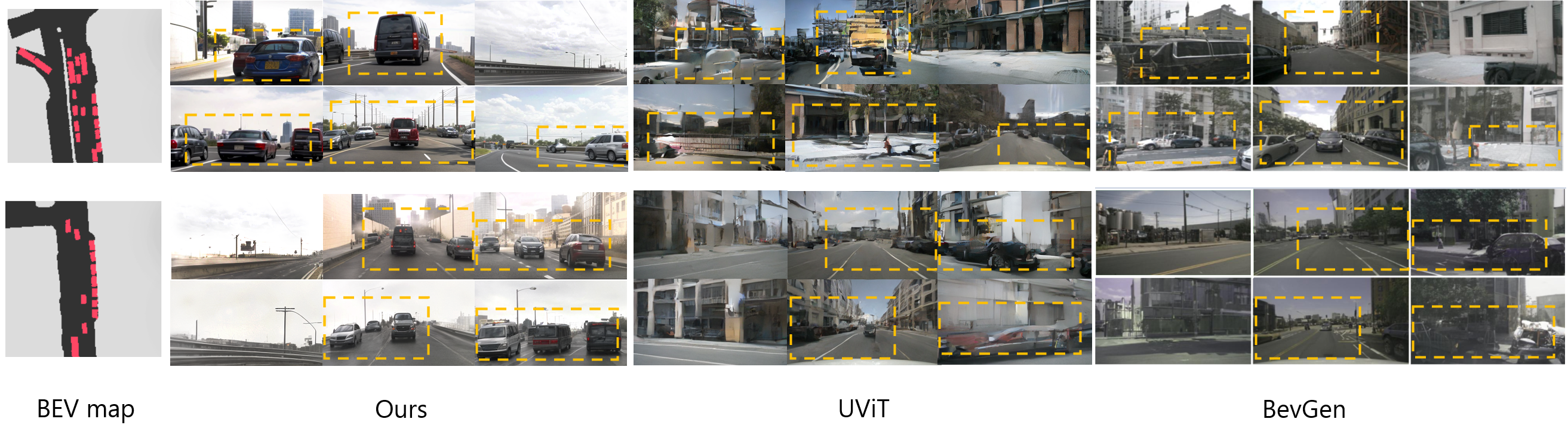}
\caption{We compare our method (left) with UViT (middle) \cite{bao2023all} and BevGen(right) \cite{swerdlow2023street}. Our results demonstrate greater stability and more effective use of conditional information, especially in the highlighted yellow regions where the condition should take effect. For best results, it is recommended to zoom in.}
\label{comp}
\end{figure*}

Informed by these insights, we fine-tune our image generator for specific viewpoints. The mechanism for view encoding is detailed in Fig. ~\ref{view}. Taking a cue from DreamBooth \cite{ruiz2023dreambooth}, which hones in on a personalized concept (e.g., a particular dog) as a unique prompt, we treat the viewpoint as an abstract concept and introduce a view-specific loss to optimize the diffusion model. This ensures that the viewpoint is distinct from foundational concepts like cars or streets within the prompts. The training loss is articulated as:

\begin{equation}
\begin{aligned}
\mathbb{E}_{\mathbf{x},\mathbf{c},\boldsymbol{\epsilon},\boldsymbol{\epsilon'},t}[w_t\|\hat{\mathbf{x}}_\theta(\alpha_t\mathbf{x}+\sigma_t\boldsymbol{\epsilon},\mathbf{c})-\mathbf{x}\|_2^2+\\
    \lambda w_{t'}\|\hat{\mathbf{x}}_\theta(\alpha_{t'}\mathbf{x}_{view}+\sigma_{t'}\boldsymbol{\epsilon'},\mathbf{c}_{view})-\mathbf{x}_{view}\|_2^2]
\end{aligned}
\end{equation}
, where $\mathbf{x}_\theta$ represents the base model, $\sigma_{t}$ and $\sigma_{t'}$ refer to distinct Gaussian noises, and $\mathbf{c}$ and $\mathbf{c}_{view}$ signify the prompt, either with or without the explicit inclusion of the viewpoint.

Rather than fine-tuning the entire network, we leverage the Low Rank Adaptation (LoRA) \cite{hu2021lora} technique to achieve rapid training and enhanced flexibility. The base model is shared across views.

\section{EXPERIMENTS AND RESULTS}

\subsection{Dataset}

The nuScenes dataset \cite{caesar2020nuscenes} is a comprehensive collection encompassing 1,000 diverse street-view scenes, captured under varied weathers, times of day, and traffic conditions. Spanning over 20 seconds, each scene consists of 40 frames, amounting to a total of 40,000 samples within the entire dataset. Designed to provide a 360° perspective around the ego-vehicle, the data is derived from six distinct camera views, capturing images from the side, front, and back of the vehicle. Every camera view comes with calibrated intrinsics (K) and extrinsics (R, t) for each timestep. Furthermore, objects, including vehicles, are consistently tracked across frames and annotated using 3D bounding boxes derived from LiDAR data. The dataset is organized into 700 training, 150 validation, and 150 testing scenes.

Following \cite{zhou2022cross}, the semantic mask of the vehicle in BEV is rendered with a resolution of (200,200). This is achieved by orthographically projecting the 3D box annotations onto the ground plane, which corresponds to a (100m,100m) region in the real-world context. The road masks are formulated using the NuScenes map devkit, which integrates both lanes and road segments.

\subsection{Implementations}

\textbf{Shape refinement network:} The shape refinement network is a convnet comprising three down-sampling blocks and four up-sampling blocks. It accepts inputs with a resolution of (56,100) and produces outputs with a resolution of (224,400). Given that the original nuScenes dataset does not include image semantic labels, we employ SegFormer \cite{xie2021segformer} to generate pseudo labels. We train the network for 10 epoches with a learning rate 1e-7.

\textbf{Pretrained Stable Diffusion and control module:} We utilize the pretrained Stable Diffusion model "RealisticVision" available on HuggingFace \cite{jain2022hugging}. The control module is adapted from ControlNet \cite{zhang2023adding}, which was originally trained on the ADE20K dataset \cite{zhou2017scene} and captioned using BLIP \cite{li2022blip}.

\textbf{Street view adaptation module:} For each camera view, we use a set of 100 images to train the respective adaptation module. Our foundational prompts for regularization include "road", "car", and "street background". To specify viewpoints, we use alphanumeric designations (e.g., cam0) to prevent any overlap with existing concepts within the pretrained CLIP text encoder \cite{radford2021learning}. During finetuning, the image resolution is set at (400, 224). The training extends over 5000 steps with a batch size of 4 and a learning rate set at 1e-4. The rank for LoRA \cite{hu2021lora} is set to 16.

\subsection{Results}

\textbf{Qualitative result:} We juxtapose our approach against BEVGen \cite{swerdlow2023street} and a from-scratch trained latent diffusion model using a transformer architecture, specifically UViT \cite{bao2023all}. Notably, our strategy involves finetuning a pre-trained, expansive model, while the other two approaches train their models from the ground up. The results can be observed in Fig. ~\ref{comp}.

Our method showcases superior stability in image quality, and its conditioning mechanism proves to be effective. Both UViT and BevGen employ cross attention to manage conditional information. However, their models occasionally falter due to the absence of explicit spatial relationships between the semantic and the resultant generated images. This makes it challenging for their conditioning mechanisms to consistently function effectively. Concerning image quality and diversity, methods that are trained from scratch tend to be closely tied to specific datasets, often risking overfitting. In particular, the UViT-based diffusion model faces challenges when trained with a limited dataset.

\begin{figure}[t]
\centering
\includegraphics[width=\linewidth]{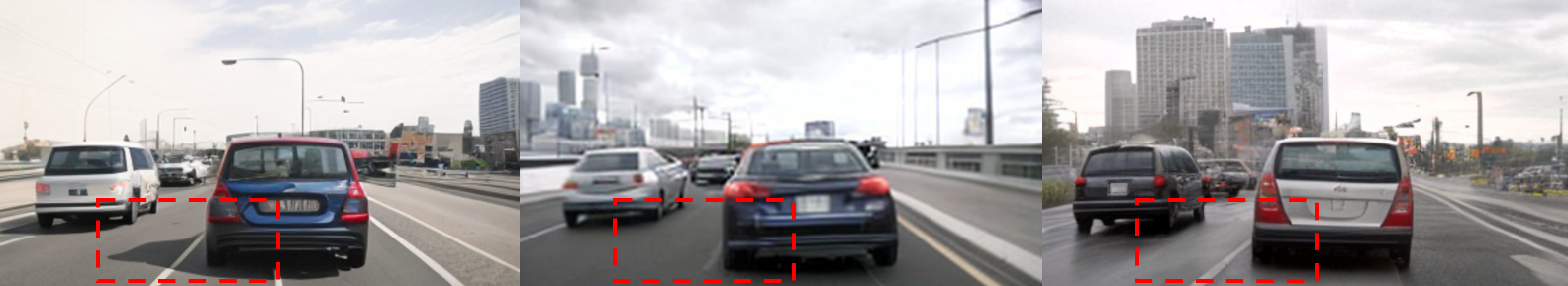}
\caption{Leveraging the robust generative capabilities of the large pretrained Stable Diffusion model, our generated outcomes display remarkable diversity. This figure presents results under varying weather conditions, all derived from a consistent BEV input. The red region illustrates the road variations in the driving scene image due to changing weather conditions. For best results, it is recommended to zoom in.}
\label{diverse}
\end{figure}

In Fig. ~\ref{diverse}, we showcase additional illustrations underscoring the diversity of our generated outcomes. Our methodology effortlessly facilitates the generation of images under various weather scenarios, significantly enhancing the model's adaptability.


\textbf{Quantitative result:} In Table. ~\ref{quan}, we juxtapose our approach with the benchmark BEVGen and a transformer-driven diffusion model. Utilizing the Frechet Inception Distance (FID) \cite{heusel2017gans}, akin to BEVGen, we evaluate the congruence between the generated images and the training dataset. While our outputs are visually appealing and consistent, our FID score lags behind BEVGen. This can be attributed to our reliance on limited data for fine-tuning, hence the visual style largely remains anchored to the foundational diffusion model. For a more equitable comparison, we trained a UViT-based latent diffusion model from scratch, which yielded an even less favorable FID score. This suggests that the scope of the training dataset might be insufficient, complicating the task of cultivating a robust diffusion model from scratch.

Further, we assessed our methodology using a pretrained BEV segmentation model \cite{zhou2022cross}. To gauge the congruity between the predicted and actual BEV segmentation maps, we employed the mean Intersection over Union (mIOU). The findings reveal that in the context of roads, our model stands shoulder to shoulder with the baseline. Given that roads are consistently obscured, it poses a challenge for our refinement model to assimilate an accurate road contour. Conversely, for vehicles, our method substantially outperforms the baseline, underscoring the potency of our segment-focused conditioning and viewpoint encoding techniques.

\begin{table}[ht]
\begin{tabular}{lccc}
\toprule
Method & FID$\downarrow$ & Road mIOU$\uparrow$ & Vehicle mIOU$\uparrow$ \\
\midrule
BEVGen \cite{swerdlow2023street} & 25.54 & 50.20 & 5.89 \\
UViT \cite{bao2023all} & 79.22 & 37.69 & 9.16 \\
Ours & 48.65 & 47.45 & 17.70 \\
\bottomrule
\end{tabular}
\caption{\label{quan} Quantitative comparision between BEVGen, UViT and our method. }
\end{table}

\subsection{Ablation Studies}

In our research, we carried out ablation studies, specifically honing in on two of our core innovations: the shape refinement process and the street view adaptation technique. The detailed results of these studies can be found in Table. ~\ref{abla}. The shape refinement process is pivotal in ensuring that map elements are accurately positioned. When the shape within the camera's perspective aligns more semantically, it resonates more effectively with the given prompt. On the other hand, the street view adaptation module plays a crucial role as a style encoder. Its primary function is to make sure that the generated images bear a strong resemblance to those in the training dataset. Moreover, this module greatly assists the image generator by enabling it to achieve proper and accurate orientations for the various map elements.

\begin{table}[ht]
\begin{tabular}{lccc}
\toprule
Method & FID$\downarrow$ & Road mIOU$\uparrow$ & Vehicle mIOU$\uparrow$ \\
\midrule
Base diffusion & 82.25 & 46.76 & 11.82 \\
+ shape refinement & 78.13 & 47.92 & 15.69 \\
+ view adaptation & 48.65 & 47.45 & 17.70 \\
\bottomrule
\end{tabular}
\caption{\label{abla} Ablation study on our core design: shape refinement and street view adaptation. }
\end{table}


\section{LIMITATIONS AND FUTURE WORKS}

\begin{figure}[t]
\centering
\includegraphics[width=\linewidth]{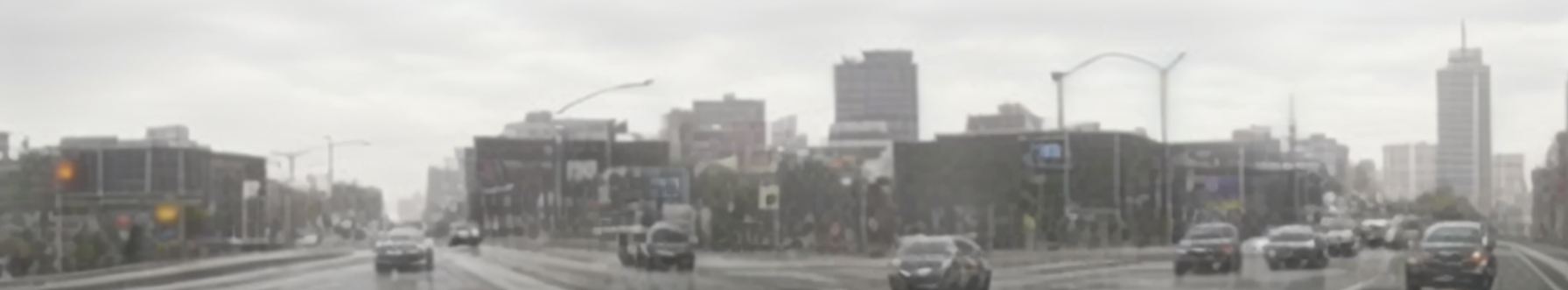}
\caption{Creating consistently aligned multi-view images poses a challenge for large pretrained diffusion models, given their typical training on standard datasets. }
\label{ring}
\end{figure}

In the specific setup we've devised, the integration of multiple cameras has the capability to produce a comprehensive panoramic image that boasts a significantly extended aspect ratio. This is a departure from traditional images and poses a unique challenge. Ideally, the most efficient approach would be to directly generate a panoramic or multi-view image, as this would inherently uphold and maintain the consistency of the view throughout the image. But herein lies the challenge: the vast majority of large-scale image diffusion models available today have been fundamentally trained to cater to standard, more conventional aspect ratios. As a result, these models, when applied to our specific need, fall short. This limitation is clearly demonstrated in Fig. ~\ref{ring}. These models face considerable difficulty when tasked with rendering high-quality images that demand a broad and expansive field-of-view.

Recognizing this gap, our future endeavors will be centered around delving deeper and exploring more robust and effective techniques that can leverage these large image diffusion models to seamlessly produce multi-view images.

\vspace{0.5cm}

\section{CONCLUSION}

We introduced an innovative framework for generating street-view images from a BEV layout by harnessing the power of a robust, pretrained latent diffusion model. Our methodology integrates view transformation, street-view adaptation, and conditional generation. When compared to baseline models trained from scratch, our model excels in terms of image quality, conditioning precision, and diversity.

\newpage

\bibliographystyle{IEEEtran}
\bibliography{reference}

\end{document}